\PassOptionsToPackage{table}{xcolor}
\documentclass[sigconf]{acmart}
\usepackage[switch]{lineno}
\usepackage{subfig}
\usepackage{amsmath}
\usepackage{longtable}
\usepackage{listings}
\usepackage{color,soul}
\usepackage{xcolor}

\DeclareFixedFont{\ttb}{T1}{txtt}{bx}{n}{12} 
\DeclareFixedFont{\ttm}{T1}{txtt}{m}{n}{12}  

\definecolor{codegreen}{rgb}{0,0.6,0}
\definecolor{codegray}{rgb}{0.5,0.5,0.5}
\definecolor{codepurple}{rgb}{0.58,0,0.82}
\definecolor{backcolour}{rgb}{0.95,0.95,0.92}

\lstdefinestyle{mystyle}{
    commentstyle=\color{codegreen},
    keywordstyle=\color{magenta},
    numberstyle=\tiny\color{codegray},
    stringstyle=\color{codepurple},
    basicstyle=\ttfamily\footnotesize,
    breakatwhitespace=false,
    breaklines=true,
    captionpos=t,
    keepspaces=true,
    frame=tb,
    numbers=left,
    numbersep=5pt,
    showspaces=false,
    showstringspaces=false,
    showtabs=false,
    tabsize=2
}

\newcommand{\listingtitle}{Example }

\newcommand{\noisetype}[1]{\textsf{#1}}
\newcommand{\noisetypei}[1]{\textsf{#1}}
\newcommand{\noisetypeb}[1]{\textbf{\textsf{#1}}}

\newcommand{\rlenv}[1]{\textit{#1}}
\newcommand{\lib}[1]{\texttt{#1}}
\newcommand{\mli}[1]{\mathit{#1}}

\newcommand{\epsgreedy}{$\epsilon\text{-greedy }$}

\renewcommand\footnotetextcopyrightpermission[1]{} 
\begin{document}
\emergencystretch 3em

\title{Simple Noisy Environment Augmentation for\\Reinforcement Learning}

\author{Raad Khraishi}
\affiliation{%
  \institution{Institute of Finance and Technology, UCL}
  \city{London}
  \country{United Kingdom}
}
\additionalaffiliation{%
  \institution{Data Science and Innovation, NatWest Group}
  \city{London}
  \country{United Kingdom}
}
\email{raad.khraishi@ucl.ac.uk}
\authornote{Corresponding author}

\author{Ramin Okhrati}
\affiliation{%
  \institution{Institute of Finance and Technology, UCL}
  \city{London}
  \country{United Kingdom}
}
\email{r.okhrati@ucl.ac.uk}

\begin{abstract}
Data augmentation is a widely used technique for improving model performance in machine learning, particularly in computer vision and natural language processing.
Recently, there has been increasing interest in applying augmentation techniques to reinforcement learning (RL) problems, with a focus on image-based augmentation.
In this paper, we explore a set of generic wrappers designed to augment RL environments with noise and encourage agent exploration and improve training data diversity which are applicable to a broad spectrum of RL algorithms and environments.
Specifically, we concentrate on augmentations concerning states, rewards, and transition dynamics and introduce two novel augmentation techniques.
In addition, we introduce a noise rate hyperparameter for control over the frequency of noise injection.
We present experimental results on the impact of these wrappers on return using three popular RL algorithms, Soft Actor-Critic (SAC), Twin Delayed DDPG (TD3), and Proximal Policy Optimization (PPO), across five MuJoCo environments.
To support the choice of augmentation technique in practice, we also present analysis that explores the performance these techniques across environments.
Lastly, we publish the wrappers in our \lib{noisyenv} repository for use with \lib{gym} environments.
\end{abstract}

\ccsdesc[500]{Theory of computation~Reinforcement learning}

\keywords{reinforcement learning, data augmentation}
\maketitle
\pagestyle{plain}

\section{Introduction}
\label{section:introduction}

Reinforcement learning (RL) has shown success in training agents to make decisions in complex environments.
However, RL algorithms often face a challenge in effectively exploring and exploiting the state space, leading to suboptimal performance. 
One potential solution to this problem is to introduce noise into the environment, such as by perturbing the state space, to encourage exploratory behavior, increase training data diversity, and improve generalization capabilities.

More generally, data augmentation is a well-established technique for improving the performance and generalization ability of machine learning algorithms in fields such as computer vision and natural language processing (NLP) \cite{wong2016understanding, perez2017effectiveness, taylor2018improving, wei2019eda}. 
Recently, there has been a growing interest in exploring augmentation methods for RL as a technique to improve exploration and enhance generalization abilities. 
While there has been significant progress in image-based augmentation as well as robotics tasks \cite{cobbe2019quantifying, lee2019network, kostrikov2020image, mezghan2022memory, laskin2020reinforcement, tobin2017domain, ren2019domain}, there has been comparatively little research on general-purpose augmentation techniques that can be applied to a diverse range of non-image-based problems, such as those found in transportation, energy management, manufacturing, marketing, and finance.

In this paper, we present a comprehensive study of the impact of different types of environment noise on the performance of standard RL algorithms. 
We focus on a range of augmentations that work across a wide-range of environments and do not require any information from or making changes to the learning algorithm or policy.
Specifically, we explore the effects of three different categories of noise: 
1) reward noise, which introduces randomness into the reward signal, 
2) state-space noise, which perturbs the state space, and 
3) dynamics noise, which introduces randomness into the transition dynamics.
We also make these wrappers available in our \lib{noisyenv} repository: \url{https://github.com/UCL-IFT/noisyenv}.

Our main contributions are threefold. 
First, we propose several novel noisy wrappers that can be used to apply different types of noise to the environment.  
Second, we extend prior work \cite{sinha2022s4rl, laskin2020reinforcement} to include noise rate as a parameter that may be tuned to control the amount of noise in the environment. 
Finally, we empirically evaluate the performance of our proposed noisy wrappers on a range of non-image-based tasks.

The remainder of this paper is structured as follows. 
Section \ref{section:related-work} provides an overview of related work.
Section \ref{section:preliminaries} introduces the necessary background concepts regarding the reinforcement learning problem.
Section \ref{section:methodology} describes our proposed noisy wrappers for environment augmentation, while Section \ref{section:results} presents our experimental results on a range of \lib{gym} tasks. 
Finally, we conclude in Section \ref{section:conclusion} with a summary of our findings including limitations and suggestions for future research.

\section{Related Work}
\label{section:related-work}

In this section, we provide an overview of related reinforcement learning (RL) literature, focusing on data augmentation, environment noise, and their applications in various RL contexts.

\paragraph{Data and states} 
There has been extensive research into image-based data augmentation for RL \cite{cobbe2019quantifying, lee2019network, kostrikov2020image, mezghan2022memory, laskin2020reinforcement}.
For example, \citet{cobbe2019quantifying} explored several types of regularization, including the cutout image augmentation technique \cite{devries2017improved}.
In DrQ, \citet{kostrikov2020image} combined simple image augmentation techniques such as random shifts with Q-function regularization that averages across one or more augmentations of a state.
In a related work, \citet{laskin2020reinforcement} presented RL with augmented data (RAD). 
Though their focus was on pixel inputs, for non-pixel inputs they evaluated the effect of adding Gaussian noise to the state and introduced the random amplitude scaling (RAS) technique, which scales the state by a random uniform value.
However, unlike our current work, they focussed on a narrow set of non-image based augmentation techniques, did not explore reward or transition noise, and did not attempt to control the noise rate.

Several recent papers have explored the usage of data augmentation in an offline setting \cite{sinha2022s4rl, joo2022swapping, lu2020sample}.
In the offline study most closely related to our present work, \citet{sinha2022s4rl} explored seven different types of state-based data augmentation applicable to non-visual problems including adding Gaussian or Uniform noise to the state and dimension dropout, where elements of the state are randomly substituted with zeroes. 
We provide further detail on some of their augmentation techniques in Section \ref{section:methodology}.

While a large focus has been on model-free methods, \citet{wang2019benchmarking} benchmarked several model-based RL algorithms on environments with added observation and action noise.
Data augmentation has also been explored in semi-supervised and contrastive learning settings \cite{park2022surf, laskin2020curl, hansen2021generalization}.
For instance, \citet{park2022surf} developed SURF, a semi-supervised reward learning framework for preference-based learning, and introduced a data augmentation technique that crops consecutive sub-sequences from the original behaviors.

Earlier studies explored simulation as a technique to enhance RL performance and generalization capabilities. 
For example, \citet{tobin2017domain} and \citet{ren2019domain} explored domain randomization to generate simulated data from a physics engine for improved Sim2Real transfer. 
Recent work \cite{kurte2022deep, liu2022synthetic} has begun to explore using a time series generative adversarial network (TimeGAN) \cite{yoon2019time} to generate synthetic trajectories.

Recent research has also explored how noise and data augmentation may be used to improve the robustness of RL algorithms to adversarial attacks \cite{mandlekar2017adversarially, antotsiou2021adversarial, zhang2020robust, zhang2021generalization, mandlekar2017adversarially}.
For example, \citet{zhang2021generalization} extended mixreg \cite{wang2020improving}, which mixes training observations from different environments, by also augmenting the training process with adversarial trajectories.

\paragraph{Rewards}
A large focus of reward noise research has been on techniques to mitigate issues stemming from corrupted or noisy reward signals.
For example, \citet{everitt2017reinforcement} defined a Corrupt Reward Markov Decision Process (CRMDP) setting where reward signals may be corrupted due to issues such as sensory errors or reward misspecification.
\citet{wang2020reinforcement} developed an RL framework to enable agents to learn in noisy environments with perturbed rewards using estimated surrogate rewards.

To encourage exploration and promote novel behavior, many RL studies have investigated intrinsic motivation and the use of reward bonuses \citep{taiga2021bonus, pathak2017curiosity, bellemare2016unifying, haarnoja2018soft}. 
For example, in the SAC \cite{haarnoja2018soft} algorithm a reward bonus proportional to the entropy of the policy is incorporated in the objective.
Novelty bonuses have also been quantified using prediction errors of a dynamics model \cite{pathak2017curiosity} or through state(-action) frequency counts \cite{bellemare2016unifying}. 

\citet{sun2021reward} explored reward space noise as a strategy for enhancing exploration and performance in RL. 
A key difference with our current work is that in their approach they introduced random normal noise perturbations to the reward signal for only the exploratory policy and used the original rewards in the learning policy, whereas, we do not modify the learning algorithm.

\paragraph{Transition dynamics}

Research into transition dynamics noise and augmentation has been quite diverse. 
\citet{szita2003varepsilon} investigated a setting in which transition probabilities are perturbed by a small noise factor over time, while \citet{padakandla2020reinforcement} explored a setting in which the underlying model of the environment changes over time.
\citet{bouteiller2021reinforcement} studied random action and environment delays, introducing Random-Delay Markov Decision Processes (RDMDP) and proposing the Delay-Correcting Actor-Critic (DCAC) algorithm to mitigate bias from delays. 
Domain randomization of environment parameters such as friction or mass has been investigated using simulators \cite{tobin2017domain, ren2019domain}.
\citet{lu2020sample} explored the use of causal models of the state dynamics for counterfactual data augmentation.
Time limits in RL have been employed to manage task complexity and facilitate learning, with \citet{pardo2018time} analyzing their utility in diversifying training experience and boosting performance when combined with agent time-awareness.

In a related work, \citet{mandlekar2017adversarially} introduced the ARPL method, which actively adds noise perturbations to state dynamics (e.g., mass and friction) and observations based on the loss function gradient, enhancing robustness against adversarial attacks.
They also experimented with random perturbations and introduced a noise rate parameter, increasing noisy perturbations during training for policy convergence. 
However, their focus was on robustness to adversarial attacks, unlike our work, which aims to enhance RL algorithm performance using noise augmentation as a general technique.

\paragraph{Actions}

Action space noise is relatively well-studied and a commonly used technique to enhance exploration.
For example, \epsgreedy \cite{watkins1989learning} and its variants are popular exploration strategies.
In addition, contemporary algorithms such as DDPG \cite{lillicrap2015continuous} and TD3 \cite{fujimoto2018addressing} add noise to actions during training time to encourage the polices to explore better.

Beyond action space noise, researchers have also investigated the use of parameter space noise and randomized value functions as a means to perturb action selection and improve exploration. 
NoisyNet \cite{fortunato2017noisy} randomly perturbs the parameters of the agent's policy network, while the approach proposed by \citet{osband2019deep} samples a value function from a distribution of value functions to determine the greedy action.

Although action space noise has been well-studied, it is outside the scope of our current work as it involves modifying the agent's output rather than the environment. 
Nonetheless, it is worth noting that the augmentation techniques explored in this paper can be used in conjunction with any of these action space noise techniques.

\section{Preliminaries}
\label{section:preliminaries}

Following the work of \citet{laskin2020reinforcement}, we consider the problem of reinforcement learning (RL) in a Markov Decision Process (MDP), defined by the tuple
$(\mathcal{S}, \mathcal{A}, P, \gamma)$,
where $\mathcal{S}$ represents the set of states $s \in \mathcal{S}$,
$\mathcal{A}$ represents the set of actions $a \in \mathcal{A}$,
$P$ represents the transition dynamics $P(s_{t+1}, r_{t} | s_{t}, a_{t})$ which define the environment's mechanics and rewards,
and $\gamma \in (0, 1)$ is the discount factor.
An RL agent interacts with the environment by selecting actions according to its policy $\pi: \mathcal{S} \rightarrow \mathcal{P}(\mathcal{A})$, where $\mathcal{P}(\mathcal{A})$ denotes the set of probability distributions over the action space. 
The agent receives rewards while transitioning between states, and its goal is to learn a policy that maximizes the return defined as the expected discounted sum of rewards, i.e., $R = \sum_{t=0}^{\infty}\gamma^{t}r_{t}$.

We also consider partial observability in which the agent potentially has restricted access to the state of the environment and instead observes an observation, $o_{t} = O(s_{t})$, where $O: \mathcal{S} \rightarrow \mathcal{O}$ is a function of the state, and $\mathcal{O}$ represents the set of observations. 
In this setting, the agent's policy depends on the observations and is represented as $\pi: \mathcal{O} \rightarrow \mathcal{P}(\mathcal{A})$. 
Note that in many environments, however, the agent may have full observability such that $o_t = s_t$ for all $t$. 

\section{Augmentation Techniques}
\label{section:methodology}

To incorporate noise in the MDPs, we explore a set of generic wrappers that perturb the state space, reward space, and transition dynamics. 
The noise rate hyperparameter, denoted as $p \in [0, 1]$, controls the probability of perturbing each step (or episode in certain cases) with noise.
In the case of reward noise, the agent observes a perturbed reward $\tilde{r}_t$, with probability $p$, 
or perturbed observation $\tilde{o}_t$ in the case of observation noise.
\listingtitle \ref{code:RandomUniformScaleReward} demonstrates an implementation of a wrapper in Python using the \lib{gym} package.

Note that these augmentation techniques can be seamlessly integrated into any environment, independent of the RL agent. 
Furthermore, they can be conveniently combined with one another or with alternative techniques such as reward clipping or action space noise to further augment the reinforcement learning process.

\subsection{State Space Noise}

\paragraph{\noisetype{RandomNormalNoisyObservation}}
This wrapper adds Gaussian noise to the observation with a probability of $p$ each step. 
The noise $\epsilon$ is sampled from $\mathcal{N}(\mathbf{0}, \sigma^{2} I)$ such that $\tilde{o}_{t} \leftarrow o_{t} + \epsilon$ where $\mathbf{0}$ is a vector of zeroes. 
Note that $\sigma > 0$ is an important hyperparameter.
When $p = 1$, this is equivalent to the Gaussian noise augmentation described by \citet{laskin2020reinforcement} and \citet{sinha2022s4rl}.

\paragraph{\noisetype{RandomUniformNoisyObservation}}
This wrapper adds uniform noise to the observation with a probability of $p$ each step such that
$\tilde{o}_{t} \leftarrow o_{t} + \epsilon$, where each dimension of $\epsilon$ is sampled from $\mathcal{U}(\alpha, \beta)$.
Both $\alpha$ and $\beta$ are important hyperparameters. 
When $p = 1$ and $\alpha = -\beta$, this is equivalent to the uniform noise augmentation described by \citet{sinha2022s4rl}.

\paragraph{\noisetype{RandomUniformScaleObservation}}
This wrapper applies random uniform noise by multiplying the observation with a probability of $p$ each step, such that 
$\tilde{o}_{t} \leftarrow o_{t} * \epsilon$, where $\epsilon$ may be either a scalar or of the same dimension as $o_{t}$, sampled from $\mathcal{U}(\alpha, \beta)$.
Both $\alpha$ and $\beta$ are important hyperparameters.
When $p = 1$, this is equivalent to the Random Amplitude Scaling (RAS) technique introduced by \citet{laskin2020reinforcement}.

\paragraph{\noisetype{RandomMixupObservation}}
This wrapper blends or replaces the current observation with the last observation with a probability of $p$ each step,  such that 
$\tilde{o}_{t} \leftarrow \lambda * o_{t} + (1 - \lambda) * o_{t - 1} $, where $\lambda \in [0, 1]$. 
Here, $\lambda$ is an important hyperparameter.
This is similar to the state mix-up augmentation technique described by \citet{sinha2022s4rl} that uses $\lambda \sim \mli{Beta}(\alpha, \alpha)$ with $\alpha=0.4$.

\paragraph{\noisetype{RandomDropoutObservation}}
This wrapper randomly replaces elements of the observation with 0 with a probability of $p$ each step, such that $\tilde{o}_{t} \leftarrow \textbf{r} * \tilde{o}_{t}$ where $\textbf{r}$ is an array of independent Bernoulli random variables each of which has a probability $\eta$ of being 1.
Here, $\eta$ is an important hyperparameter.
This can be considered a general version of dimension dropout described by \citet{sinha2022s4rl}.

\subsection{Reward Space Noise}

\paragraph{\noisetype{RandomNormalNoisyReward}}
This wrapper adds zero-mean Gaussian noise to the reward, such that 
$\tilde{r}_{t} \leftarrow r_{t} + \epsilon$, where $\epsilon$ is sampled from $\mathcal{N}(0, \sigma ^2)$.
Here, $\sigma$ is an important hyperparameter.
\citet{sun2021reward} explored a similar perturbation of the reward signal, however, they also incorporated the unperturbed reward in their algorithm. 

\paragraph{\noisetype{RandomUniformNoisyReward}}
This wrapper adds uniform noise to the reward, such that 
$\tilde{r}_{t} \leftarrow r_{t} + \epsilon$, where $\epsilon$ is sampled from $\mathcal{U}(\alpha, \beta)$.
Both $\alpha$ and $\beta$ are important hyperparameters.

\paragraph{\noisetype{RandomUniformScaleReward}}
This wrapper scales the reward by a random uniform factor, such that, 
$\tilde{r}_{t} \leftarrow r_{t} * \epsilon$, where $\epsilon$ is sampled from $\mathcal{U}(\alpha, \beta)$.
Both $\alpha$ and $\beta$ are important hyperparameters.
To the best of our knowledge, this augmentation technique is novel.
An implementation of this technique is shown in \listingtitle \ref{code:RandomUniformScaleReward}.

\begin{lstlisting}[
    language=Python,
    label=code:RandomUniformScaleReward,
    caption=Example Python implementation of the \noisetype{RandomUniformScaleReward} wrapper,
    float=htpb
]
class RandomUniformScaleReward(gym.RewardWrapper):
    def init(self, env, noise_rate, low, high):
        super().init(env)
        self.noise_rate = noise_rate
        self.low = low
        self.high = high

    def reward(self, reward):
        if np.random.rand() < self.noise_rate:
            scale_factor = np.random.uniform(
                self.low, self.high
            )
            return reward * scale_factor
        else:
            return reward
\end{lstlisting}

\subsection{Transition Dynamics Noise}

\paragraph{\noisetype{RandomEarlyTermination}}
This wrapper randomly chooses an episode with probability $p$ and terminates it early at a specific step.
The termination step is a randomly sampled integer from the interval $[a, b]$, where $a$ and $b$ are integers and important hyperparameters. 
An alternative version of this augmentation uses $p$ as the probability any individual step terminates early which removes the need for a specific termination step.
To the best of our knowledge, this augmentation technique is novel.

\section{Experimental Results}
\label{section:results}

\begin{table*}[ht!]
    \caption{Final average episodic returns and standard deviations after 1M steps for the best noise augmentation strategy (Noisy) for each environment and algorithm versus baseline (Baseline) performance with no noise added to the environment during training. Results are obtained using a separate test set without any added noise averaged over 10 seeds for PPO and 5 seeds for SAC and TD3. See Appendix \ref{appendix:results} for full results.}
    \resizebox{\linewidth}{!}{%
        \begin{tabular}{lllcrrr}
\toprule
    &             &                                     Noise Type &  Noise Rate ($p$) &               Baseline &                  Noisy &  \% Imp. \\
Algorithm & Environment &                                                &                   &                        &                        &          \\
\midrule
PPO & HalfCheetah-v2 &       RandomUniformNoisyReward (-0.001, 0.001) &              1.00 &    1,685.8 $\pm$ 653.4 &    2,293.6 $\pm$ 966.3 &     36.1 \\
    & Hopper-v2 &                  RandomNormalNoisyReward (1.0) &              1.00 &    1,469.5 $\pm$ 786.4 &  1,878.4 $\pm$ 1,279.8 &     27.8 \\
    & Humanoid-v2 &                  RandomNormalNoisyReward (1.0) &              0.01 &      753.7 $\pm$ 255.8 &      779.5 $\pm$ 163.9 &      3.4 \\
    & Pusher-v2 &                    RandomEarlyTermination (50) &              1.00 &       -57.6 $\pm$ 14.1 &        -45.9 $\pm$ 9.1 &     20.4 \\
    & Walker2d-v2 &            RandomUniformScaleReward (0.8, 1.2) &              0.05 &  1,806.0 $\pm$ 1,206.7 &  3,424.6 $\pm$ 1,481.3 &     89.6 \\
SAC & HalfCheetah-v2 &       RandomUniformNoisyReward (-0.001, 0.001) &              1.00 &  9,515.4 $\pm$ 1,706.8 &   11,772.0 $\pm$ 259.9 &     23.7 \\
    & Hopper-v2 &                   RandomMixupObservation (0.5) &              0.05 &    2,928.0 $\pm$ 957.0 &    3,597.2 $\pm$ 164.6 &     22.9 \\
    & Humanoid-v2 &           RandomNormalNoisyObservation (0.001) &              1.00 &  4,071.9 $\pm$ 1,913.1 &    5,463.8 $\pm$ 129.0 &     34.2 \\
    & Pusher-v2 &       RandomUniformNoisyReward (-0.001, 0.001) &              0.50 &        -22.7 $\pm$ 1.8 &        -20.8 $\pm$ 1.5 &      8.3 \\
    & Walker2d-v2 &  RandomUniformNoisyObservation (-0.001, 0.001) &              1.00 &    4,767.0 $\pm$ 440.6 &    5,372.8 $\pm$ 434.0 &     12.7 \\
TD3 & HalfCheetah-v2 &  RandomUniformNoisyObservation (-0.001, 0.001) &              0.50 &   10,123.6 $\pm$ 396.5 &   10,206.8 $\pm$ 474.1 &      0.8 \\
    & Hopper-v2 &       RandomUniformScaleObservation (0.5, 1.5) &              0.05 &    2,934.8 $\pm$ 906.2 &    3,499.5 $\pm$ 151.7 &     19.2 \\
    & Humanoid-v2 &       RandomUniformScaleObservation (0.8, 1.2) &              0.20 &  4,800.6 $\pm$ 1,258.0 &    5,360.7 $\pm$ 159.9 &     11.7 \\
    & Pusher-v2 &  RandomUniformNoisyObservation (-0.001, 0.001) &              0.20 &        -29.6 $\pm$ 4.8 &        -27.3 $\pm$ 1.7 &      7.8 \\
    & Walker2d-v2 &  RandomUniformNoisyObservation (-0.001, 0.001) &              0.50 &    4,015.6 $\pm$ 415.4 &    4,502.8 $\pm$ 601.7 &     12.1 \\
\bottomrule
\end{tabular}

    }
  \label{table:results-best}
\end{table*}

\subsection{Setup} 
We evaluated the performance of the data augmentation techniques described in Section \ref{section:methodology} using three popular model-free off-policy and on-policy RL algorithms: Soft Actor-Critic (SAC) \cite{haarnoja2018soft}, Twin Delayed DDPG (TD3) \cite{fujimoto2018addressing}, and Proximal Policy Optimization (PPO) \cite{schulman2017proximal}. 
The implementations of these algorithms were obtained from the \lib{CleanRL} package \cite{huang2022cleanrl} with minor modifications to create separate training and testing environments and to allow for wrapping of the training environment with the noisy augmentation techniques.
Note that for evaluation the test environment remained unaugmented without any added noise in each of the experiments.
We used the default hyperparameters for each of the algorithms in \lib{CleanRL} which we list in Appendix \ref{appendix:reproducibility}.

The algorithms were applied to five different continuous control MuJoCo OpenAI gym environments \cite{brockman2016gym} with varying levels of complexity, including \rlenv{HalfCheetah-v2}, \rlenv{Hopper-v2}, \rlenv{Humanoid-v2}, \rlenv{Pusher-v2}, and \rlenv{Walker2d-v2}.\footnote{We used v2 of these environments to compare against \lib{CleanRL}'s public benchmarks.}
See Appendix \ref{appendix:envs} for more details on the environments.

For each noise augmentation, we evaluated the performance across six different noise rates, $p \in \{ 0.01, 0.05, 0.10, 0.20, 0.50, 1.0 \}$.
We include information on the hyperparameters used for the data augmentation techniques in Appendix \ref{appendix:noisy-wrappers}.

\begin{figure*}[ht!]%
    \centering
    \subfloat[\centering HalfCheetah-v2]{{\includegraphics[width=0.3\linewidth]{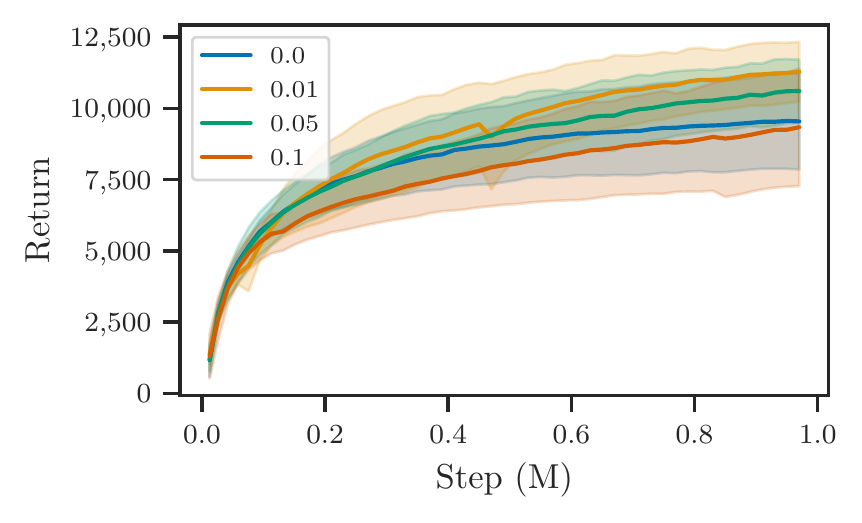} }}%
    \qquad
    \subfloat[\centering Hopper-v2]{{\includegraphics[width=0.3\linewidth]{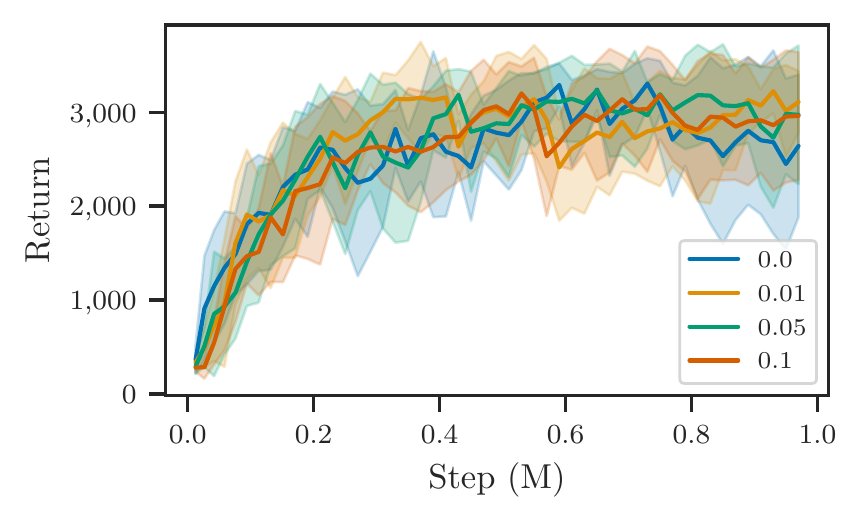} }}%
    \qquad
    \subfloat[\centering Humanoid-v2]{{\includegraphics[width=0.3\linewidth]{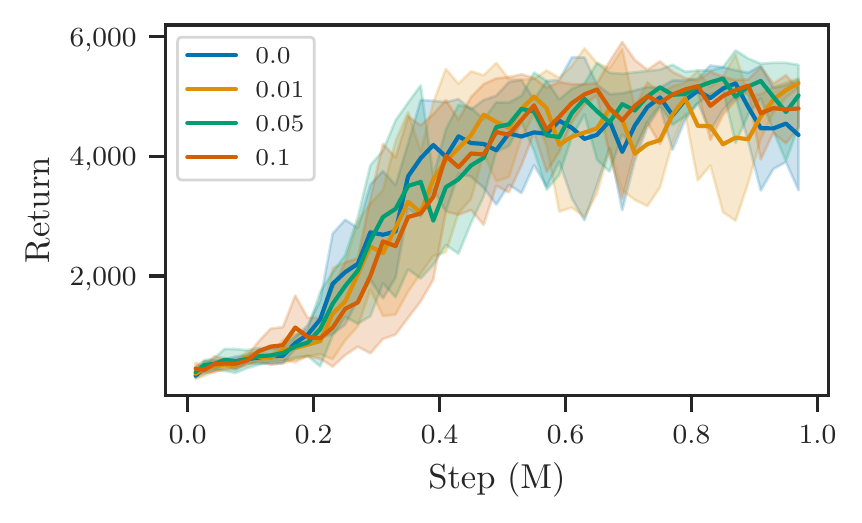} }}%
    \qquad
    \subfloat[\centering Walker2d-v2]{{\includegraphics[width=0.3\linewidth]{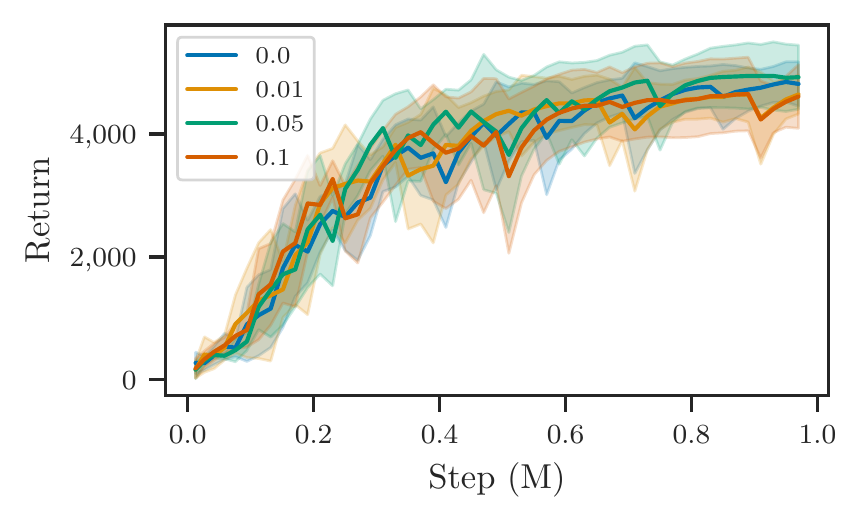} }}%
    \qquad
    \subfloat[\centering Pusher-v2]{{\includegraphics[width=0.3\linewidth]{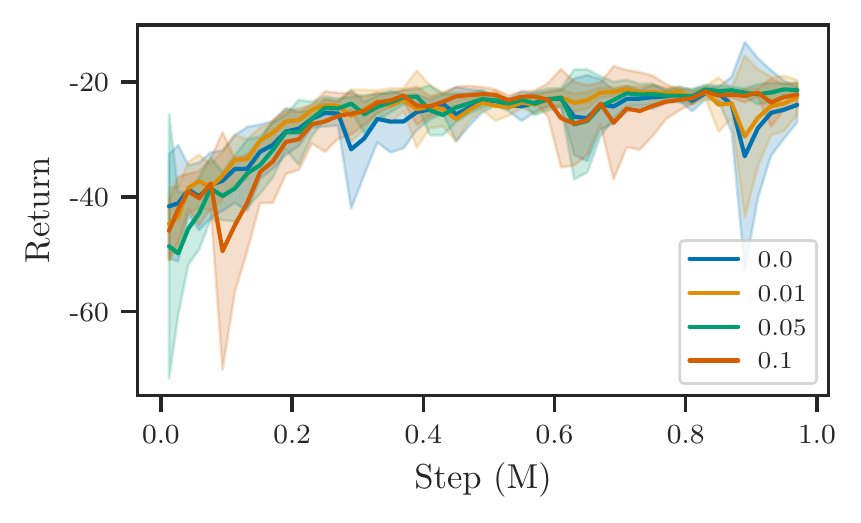} }}%
    \caption{Average episodic return on the unaugmented test environments across five seeds using the \noisetype{RandomUniformScaleReward (0.5, 1.5)} wrapper for the SAC algorithm with each line representing a different noise rate and the shaded region representing the standard deviation across seeds.}
    \label{fig:learning-curves}%
\end{figure*}

\subsection{Analysis} 

Our results provide initial evidence that specific noise augmentation techniques may potentially be used to enhance the performance of RL algorithms. 
For example, in Table \ref{table:results-best} we showcase the top performing augmentation technique in terms of return for each algorithm and environment and compare against a baseline that did not use noise augmentation during training.
The potential performance improvements we observe in these best-case examples are substantial, with a maximum increase of approximately 90\% for PPO, 19\% for TD3, and 34\% for SAC averaged across seeds.\footnote{We note that the nature of the environments and algorithms we dealt with in our study inherently introduces variability, leading to statistical uncertainty that is common in most RL research which may make it difficult to draw firm conclusions \cite{agarwal2021deep}.}
Moreover, we identify only a single instance in which none of the noise augmentation techniques could yield a noticeable enhancement in the algorithm's performance: TD3 applied to the \rlenv{HalfCheetah-v2} environment.
A comprehensive list of our results, including instances where noise augmentation did not enhance performance, can be found in Appendix \ref{appendix:results}.

In addition to achieving higher returns through noise augmentation, we also observe a possible reduction in performance variance in some cases.
For example, when employing \noisetype{RandomUniformNoisyReward (-0.001, 0.001)} with a noise rate, $p$, of 1.0 with the SAC algorithm on the \rlenv{HalfCheetah-v2} environment, we observe not only a performance improvement of 23.7\% but also a reduction in performance variance of over 80\%.
Furthermore, we find preliminary evidence of enhanced sample efficiency and accelerated performance with some augmentation techniques.
As illustrated in Figure \ref{fig:learning-curves}, utilizing \noisetype{RandomUniformScaleReward (0.5, 1.5)} with the SAC algorithm and a noise rate of 0.01, surpasses the baseline's final return in less than half the steps.
We see further evidence of this in Appendix \ref{appendix:results}, Table \ref{table:results-early-performance} which includes a summary of early performance for SAC with the \noisetype{RandomUniformScaleReward (0.5, 1.5)} augmentation.

\begin{table*}[ht!]
    \caption{Noise augmentations yielding consistent performance (improved performance in 80\% or more of the environments) for each algorithm. The \textit{\% Env.} column denotes the percentage of environments in which the augmentation made an improvement, and the \textit{Avg. \% Imp.} column displays the average percentage improvement over the unaugmented baseline and standard deviation across environments.}
    \begin{tabular}{llcrr}
\toprule
    &                                          &      &  \% of Envs. &     Avg. \% Imp. \\
Algorithm & Noise Type & Noise Rate ($p$) &              &                  \\
\midrule
TD3 & RandomUniformNoisyObservation (-0.001, 0.001) & 0.50 &          100 &    5.6 $\pm$ 5.3 \\
    & RandomNormalNoisyObservation (0.001) & 0.05 &          100 &    4.3 $\pm$ 6.3 \\
    & RandomMixupObservation (0.0) & 0.01 &           80 &    4.9 $\pm$ 4.9 \\
    & RandomUniformScaleReward (0.5, 1.5) & 0.05 &           80 &    4.3 $\pm$ 9.5 \\
SAC & RandomNormalNoisyReward (0.001) & 0.05 &          100 &  15.7 $\pm$ 11.9 \\
    & RandomUniformScaleReward (0.5, 1.5) & 0.05 &          100 &  10.9 $\pm$ 13.1 \\
    & RandomMixupObservation (0.5) & 0.05 &          100 &    8.0 $\pm$ 8.7 \\
    & RandomNormalNoisyObservation (0.001) & 0.10 &           80 &  14.6 $\pm$ 14.1 \\
    & RandomUniformScaleReward (0.5, 1.5) & 0.01 &           80 &  14.0 $\pm$ 12.4 \\
    & RandomUniformScaleReward (0.8, 1.2) & 0.01 &           80 &  12.9 $\pm$ 12.1 \\
    & RandomUniformNoisyObservation (-0.001, 0.001) & 0.05 &           80 &  11.9 $\pm$ 12.8 \\
    &                                          & 0.10 &           80 &  10.9 $\pm$ 10.6 \\
    &                                          & 1.00 &           80 &   9.0 $\pm$ 11.4 \\
    & RandomNormalNoisyReward (0.001) & 0.20 &           80 &   8.0 $\pm$ 15.0 \\
    & RandomNormalNoisyObservation (0.001) & 0.20 &           80 &   7.1 $\pm$ 16.3 \\
    & RandomNormalNoisyReward (0.001) & 1.00 &           80 &   7.0 $\pm$ 14.3 \\
    & RandomUniformNoisyReward (-0.001, 0.001) & 0.10 &           80 &   6.4 $\pm$ 19.5 \\
    & RandomUniformNoisyObservation (-0.001, 0.001) & 0.20 &           80 &   5.5 $\pm$ 11.7 \\
    & RandomUniformScaleReward (0.8, 1.2) & 0.50 &           80 &    5.2 $\pm$ 8.5 \\
    & RandomUniformNoisyObservation (-0.001, 0.001) & 0.50 &           80 &   4.8 $\pm$ 23.0 \\
    & RandomUniformScaleObservation (0.5, 1.5) & 0.01 &           80 &    4.7 $\pm$ 6.2 \\
    & RandomDropoutObservation (0.1) & 0.01 &           80 &   3.9 $\pm$ 17.7 \\
    & RandomUniformScaleReward (0.5, 1.5) & 0.20 &           80 &   3.2 $\pm$ 22.0 \\
    & RandomMixupObservation (0.5) & 0.10 &           80 &    2.3 $\pm$ 4.8 \\
PPO & RandomUniformScaleReward (0.8, 1.2) & 0.05 &           80 &  20.3 $\pm$ 39.8 \\
    & RandomUniformNoisyObservation (-0.001, 0.001) & 1.00 &           80 &   8.2 $\pm$ 19.3 \\
    & RandomNormalNoisyReward (0.001) & 0.10 &           80 &   4.1 $\pm$ 17.7 \\
    & RandomUniformScaleReward (0.5, 1.5) & 0.01 &           80 &   2.3 $\pm$ 12.2 \\
    & RandomUniformScaleObservation (0.5, 1.5) & 0.10 &           80 &    0.6 $\pm$ 7.8 \\
\bottomrule
\end{tabular}

  \label{table:results-count-imp}
\end{table*}

To support the selection of noise augmentation techniques in practice, we provide initial analysis exploring which techniques demonstrate consistent performance across multiple environments. 
Table \ref{table:results-count-imp} presents the augmentation techniques that matched or outperformed the baseline (i.e., without training augmentation) in at least four out of the five environments from which we identify several augmentations potentially capable of enhancing TD3 and SAC performance on average across all environments. 
For example, with a noise rate, $p$, of 0.05, \noisetype{RandomNormalNoisyReward (0.001)}, \noisetype{RandomUniformScaleReward (0.5, 1.5)}, and \noisetype{RandomMixupObservation (0.5)} improve SAC performance by 15.7\%, 10.9\%, and 8.0\%, respectively, across all five environments on average.
Nevertheless, it is important to note that the augmentation techniques offering the most consistent performance across environments may not necessarily yield the best overall performance as we see comparing Tables \ref{table:results-best} and \ref{table:results-count-imp}.

In addition to identifying effective augmentation techniques, we also investigated which approaches consistently worsened performance, as detailed in Appendix \ref{appendix:performance}, Table \ref{table:results-count-bad}. 
Our results demonstrate the sensitivity of the algorithm to the choice of noise type, and we observe that certain noise types and rates can lead to drastically decreased performance for specific algorithms.
We also find several examples in which certain noise types fail across all noise rates, for example, \noisetype{RandomNormalNoisyObservation (1.0)} with TD3.

Overall, our results underscore the importance of noise type as a critical choice. 
Specifically, the effectiveness of certain noise types appears to be algorithm-dependent, with different noise types exhibiting superior performance across various algorithms. 
This is exemplified by PPO's inclination towards reward noise in four out of the top five noise wrappers across various environments in Table \ref{table:results-best}, in contrast to TD3's preference for observation noise across all.
In Appendix \ref{appendix:ablations} we provide initial studies exploring the effect of augmentation technique hyperparameters on performance for the PPO algorithm which demonstrate the sensitivity of the algorithm performance to the scale of added noise.

We also find that the noise rate, $p$, is a vital hyperparameter, and its effect on performance also varies by algorithm and augmentation technique, as detailed in Appendix \ref{appendix:results}. 
In particular, we observe that a noise rate of 1.0 does not consistently yield optimal results, as demonstrated by only a handful of instances of top performance at this rate in Table \ref{table:results-best} and Table \ref{table:results-count-imp}.
This observation holds true even for several examples utilizing the augmentation techniques proposed by \citet{laskin2020reinforcement} and \citet{sinha2022s4rl}, where the authors did not originally control the noise rate.

In addition to potential improvements related to the introduction of the noise rate parameter, we also observe initial evidence of promising performance for the \noisetype{RandomUniformScaleReward} technique introduced in this paper in Table \ref{table:results-count-imp}.
Notably, \noisetype{RandomUniformScaleReward (0.5, 1.5)} seems to offer consistent performance for all three algorithms, reinforcing its potential as a versatile augmentation technique for a broad range of reinforcement learning tasks.
While the other technique introduced in this paper, \noisetype{RandomEarlyTermination}, does not produce similar results, we identify several instances in Appendix \ref{appendix:results} where it surpasses the baseline performance. 
Our experiments also indicate that this augmentation technique presents greater tuning challenges, primarily due to the high sensitivity to the choice of its hyperparameters $a$ and $b$.

\section{Discussion}
\label{section:conclusion}

\paragraph{Summary}
In this paper, we presented a comprehensive investigation of noisy environment augmentation as a technique for enhancing the exploration, robustness, and generalization capabilities of reinforcement learning algorithms. 
We proposed a set of generic wrappers that can be used to augment RL environments with different types of noise, including reward noise, state-space noise, and transition dynamics noise.
These wrappers are designed to be easily applicable to a wide range of RL algorithms and environments.
Through extensive experiments across five MuJoCo environments using three popular RL algorithms, we demonstrated that the introduction of specific types of noise into the environment may be used to improve the performance of these algorithms. 
Additionally, we discovered that adjusting the noise rate may help to fine-tune the effects of noisy environment augmentation.
We also provided initial analysis on the performance of these techniques, identifying those that consistently yield improved or diminished results.

A plausible explanation for the effectiveness of noise augmentation in certain cases is the increased diversity of training data and complexity of the environment. 
By introducing noise, the agent is compelled to adapt to fluctuating conditions, potentially leading to improved exploration and robustness.

\paragraph{Limitations and Future Work}

It is important to acknowledge that the nature of the environments and policies we dealt with in our study inherently introduces variability, leading to statistical uncertainty that is common in most RL research.
In addition, our approach has some limitations, particularly concerning the selection of noise type and noise rate employed to augment the training environment. 
While in certain cases, a specific noise augmentation may improve performance across environments, this is not consistently observed, and the optimal choice is not always apparent. 
A potential direction for future research is to investigate the automation of noise augmentation selection, drawing inspiration from techniques such as DrAC \cite{raileanu2021automatic} and AutoAugment \cite{cubuk2019autoaugment}.

Another limitation arises from the instability introduced by certain noise augmentations. 
For instance, there are several cases in which performance variance increases substantially, and at least one instance where no noise augmentation leads to improved performance. 
As part of our future work, we aim to explore the effects of combining different noise types and develop adaptive methods to adjust the noise rate during training, which may help address these limitations and further refine the efficacy of noisy environment augmentation in reinforcement learning tasks.

\begin{acks}
Raad Khraishi's research is currently funded by NatWest Group.
We thank Greig Cowan, Graham Smith, and Zachery Anderson for their valuable feedback and support.
We would also like to thank Salvatore Mercuri for his feedback on earlier drafts.
\end{acks}

\bibliographystyle{ACM-Reference-Format}
\bibliography{citations}

\appendix

\vfill\eject
\section{Appendix}
\label{appendix:appendix1}

\subsection{Environments}
\label{appendix:envs}

Table \ref{table:env-description} provides more details on the dimensionality of the environments used.

\begin{table}[h!]
    \caption{Observation dimensions, action dimensions, and horizons for each of the MuJoCo environments used.}
    \begin{tabular}{lrrr}
\toprule
   Environment &  Observation Dim. &  Action Dim. &  Horizon \\
\midrule
HalfCheetah-v2 &                17 &            6 &     1000 \\
     Hopper-v2 &                11 &            3 &     1000 \\
   Humanoid-v2 &               376 &           17 &     1000 \\
     Pusher-v2 &                23 &            7 &      100 \\
   Walker2d-v2 &                17 &            6 &     1000 \\
\bottomrule
\end{tabular}

  \label{table:env-description}
\end{table}

\subsection{Experimental Setup and Reproducibility}
\label{appendix:reproducibility}

We utilized the default implementations of the SAC, TD3, and PPO algorithms from the \lib{CleanRL} package \cite{huang2022cleanrl}, making minor modifications to accommodate our augmentation wrappers and noise rate parameter, as well as to enable a separate unaugmented test environment. 
The default hyperparameters employed for each of the algorithms are provided in Table \ref{table:algo-hyperparams}.

In order to ensure a robust evaluation of our approach, we used the benchmark utility supplied within the \lib{CleanRL} package. 
We conducted experiments with ten different seeds for PPO and five seeds for both SAC and TD3, starting with a seed value of one, across the MuJoCo environments.

Our experiments were conducted on an AMD Ryzen Threadripper PRO CPU, which facilitated the parallel distribution of runs across multiple threads.
For the SAC algorithm, a single run took approximately one and a half days to complete with the \rlenv{Humanoid-v2} environment, while the other MuJoCo environments required approximately half a day per run.
In the case of TD3, the \rlenv{Humanoid-v2} environment required slightly over a day for a single run, and the remaining MuJoCo environments took less than half a day per run.
Lastly, the PPO algorithm completed a single run with the \rlenv{Humanoid-v2} environment in under two hours, while the other MuJoCo environments were executed in less than one hour.

\begin{table}[ht!]
    \caption{Relevant \lib{CleanRL} hyperparameters used for SAC, TD3, and PPO.}
    \begin{tabular}{lrrr}
\toprule
{} &      SAC &      TD3 &     PPO \\
\midrule
alpha                    &      0.2 &          &         \\
autotune                 &     True &          &         \\
batch\_size               &      256 &      256 &    2048 \\
buffer\_size              &  1000000 &  1000000 &         \\
gamma                    &     0.99 &     0.99 &    0.99 \\
learning\_starts          &     5000 &    25000 &         \\
policy\_frequency         &        2 &        2 &         \\
policy\_lr                &   0.0003 &          &         \\
q\_lr                     &    0.001 &          &         \\
target\_network\_frequency &        1 &          &         \\
tau                      &    0.005 &    0.005 &         \\
exploration\_noise        &          &      0.1 &         \\
learning\_rate            &          &   0.0003 &  0.0003 \\
noise\_clip               &          &      0.5 &         \\
policy\_noise             &          &      0.2 &         \\
anneal\_lr                &          &          &    True \\
clip\_coef                &          &          &     0.2 \\
clip\_vloss               &          &          &    True \\
ent\_coef                 &          &          &       0 \\
gae\_lambda               &          &          &    0.95 \\
max\_grad\_norm            &          &          &     0.5 \\
minibatch\_size           &          &          &      64 \\
norm\_adv                 &          &          &    True \\
num\_minibatches          &          &          &      32 \\
num\_steps                &          &          &    2048 \\
update\_epochs            &          &          &      10 \\
vf\_coef                  &          &          &     0.5 \\
\bottomrule
\end{tabular}

  \label{table:algo-hyperparams}
\end{table}

\subsection{Noisy Wrappers}
\label{appendix:noisy-wrappers}

\begin{itemize}
\item \noisetypeb{RandomDropoutObservation (0.1):}
This wrapper applies the \noisetypei{RandomDropoutObservation} augmentation with $\eta=0.1$.

\item \noisetypeb{RandomEarlyTermination (1, 100):}
This wrapper applies the \noisetypei{RandomEarlyTermination} augmentation with $a=1$ and $b=100$.

\item \noisetypeb{RandomEarlyTermination (50):} 
This wrapper applies the \noisetypei{RandomEarlyTermination} augmentation with $a=50$ and $b=50$.

\item \noisetypeb{RandomMixupObservation (0.0):}
This wrapper applies the \noisetypei{RandomMixupObservation} augmentation with $\lambda=0.0$.

\item \noisetypeb{RandomMixupObservation (0.5):}
This wrapper applies the \noisetypei{RandomMixupObservation} augmentation with $\lambda=0.5$.

\item \noisetypeb{RandomNormalNoisyObservation (0.001):}
This wrapper applies the  \noisetypei{RandomNormalNoisyObservation} with $\sigma=0.001$.

\item \noisetypeb{RandomNormalNoisyReward (1.0):}
This wrapper applies the  \noisetypei{RandomNormalNoisyReward} with $\sigma=1.0$.

\item \noisetypeb{RandomNormalNoisyReward (0.001):}
This wrapper applies the  \noisetypei{RandomNormalNoisyReward} with $\sigma=0.001$.

\item \noisetypeb{RandomUniformNoisyObservation (-0.001, 0.001):}
This wrapper applies the  \noisetypei{RandomUniformNoisyObservation} with $\alpha=-0.001$ and $\beta=0.001$.

\item \noisetypeb{RandomUniformNoisyReward (-0.001, 0.001):}
This wrapper applies the  \noisetypei{RandomUniformNoisyReward} with $\alpha=-0.001$ and $\beta=0.001$.

\item \noisetypeb{RandomUniformScaleObservation (0.5, 1.5):}
This wrapper applies the  \noisetypei{RandomUniformScaleObservation} with $\alpha=0.5$ and $\beta=1.5$.

\item \noisetypeb{RandomUniformScaleObservation (0.8, 1.2):}
This wrapper applies the  \noisetypei{RandomUniformScaleObservation} with $\alpha=0.8$ and $\beta=1.2$.

\item \noisetypeb{RandomUniformScaleReward (0.5, 1.5):}
This wrapper applies the  \noisetypei{RandomUniformScaleReward} with $\alpha=0.5$ and $\beta=1.5$.

\item \noisetypeb{RandomUniformScaleReward (0.8, 1.2):}
This wrapper applies the  \noisetypei{RandomUniformScaleReward} with $\alpha=0.8$ and $\beta=1.2$.

\end{itemize}

\subsection{Performance Analysis}
\label{appendix:performance}

Table \ref{table:results-early-performance} includes early performance results for the SAC algorithm using the \noisetype{RandomUniformScaleReward (0.5, 1.5)} augmentation technique.
Table \ref{table:results-count-bad} contains a summary of the noise augmentation that consistently performed poorly across the various environments for each algorithm.

\begin{table}[h!]
    \caption{Early performance of the SAC algorithm on the unaugmented test environment using the \noisetype{RandomUniformScaleReward (0.5, 1.5)} technique noise rates of 0.01 and 0.05 compared to baseline performance with no training augmentation (0.0) averaged across five seeds. The column \textit{Steps (M)} denotes the number of training steps (in millions) before evaluation.}
    \resizebox{\linewidth}{!}{%
    \begin{tabular}{rrrrr}
\toprule
            &     &               0.00 &                0.01 &                0.05 \\
Environment & Steps (M) &                    &                     &                     \\
\midrule
HalfCheetah-v2 & 0.2 &    7,295 $\pm$ 899 &   7,421 $\pm$ 1,331 &     7,205 $\pm$ 891 \\
            & 0.4 &  8,461 $\pm$ 1,295 &   9,145 $\pm$ 1,400 &   8,784 $\pm$ 1,217 \\
            & 0.6 &  9,047 $\pm$ 1,512 &  10,229 $\pm$ 1,299 &   9,512 $\pm$ 1,185 \\
            & 0.8 &  9,322 $\pm$ 1,639 &    10,876 $\pm$ 987 &  10,300 $\pm$ 1,088 \\
            & 1.0 &  9,515 $\pm$ 1,707 &  11,334 $\pm$ 1,060 &  10,620 $\pm$ 1,072 \\
Hopper-v2 & 0.2 &    2,529 $\pm$ 605 &   2,023 $\pm$ 1,195 &   2,619 $\pm$ 1,003 \\
            & 0.4 &    3,166 $\pm$ 697 &     3,182 $\pm$ 888 &     2,863 $\pm$ 980 \\
            & 0.6 &    3,230 $\pm$ 433 &     2,580 $\pm$ 945 &     3,247 $\pm$ 754 \\
            & 0.8 &  2,173 $\pm$ 1,320 &     2,998 $\pm$ 584 &     2,764 $\pm$ 775 \\
            & 1.0 &    2,928 $\pm$ 957 &     3,255 $\pm$ 785 &     2,959 $\pm$ 912 \\
Humanoid-v2 & 0.2 &    1,203 $\pm$ 335 &       843 $\pm$ 361 &       989 $\pm$ 347 \\
            & 0.4 &  4,420 $\pm$ 1,734 &   2,793 $\pm$ 1,569 &   3,673 $\pm$ 2,074 \\
            & 0.6 &  4,338 $\pm$ 1,504 &   4,507 $\pm$ 1,605 &   3,754 $\pm$ 1,759 \\
            & 0.8 &  4,245 $\pm$ 1,647 &      5,345 $\pm$ 85 &   4,841 $\pm$ 1,207 \\
            & 1.0 &  4,072 $\pm$ 1,913 &      5,394 $\pm$ 68 &     5,425 $\pm$ 273 \\
Pusher-v2 & 0.2 &        -29 $\pm$ 9 &         -27 $\pm$ 3 &         -28 $\pm$ 6 \\
            & 0.4 &       -27 $\pm$ 10 &         -21 $\pm$ 2 &         -21 $\pm$ 1 \\
            & 0.6 &        -23 $\pm$ 1 &         -24 $\pm$ 3 &         -25 $\pm$ 5 \\
            & 0.8 &        -22 $\pm$ 2 &         -22 $\pm$ 2 &         -23 $\pm$ 2 \\
            & 1.0 &        -23 $\pm$ 2 &         -21 $\pm$ 1 &         -22 $\pm$ 2 \\
Walker2d-v2 & 0.2 &  2,750 $\pm$ 1,347 &   2,860 $\pm$ 1,373 &   1,913 $\pm$ 1,556 \\
            & 0.4 &  3,365 $\pm$ 1,433 &   3,583 $\pm$ 2,051 &   3,586 $\pm$ 1,872 \\
            & 0.6 &  3,317 $\pm$ 2,054 &     4,538 $\pm$ 346 &     4,850 $\pm$ 545 \\
            & 0.8 &    4,808 $\pm$ 353 &     4,579 $\pm$ 412 &     4,957 $\pm$ 550 \\
            & 1.0 &    4,767 $\pm$ 441 &     4,766 $\pm$ 418 &     4,940 $\pm$ 490 \\
\bottomrule
\end{tabular}

}
  \label{table:results-early-performance}
\end{table}

\begin{table*}[ht!]
    \caption{Noise augmentations yielding consistently poor performance (decreased performance in all of the environments) for each algorithm. The \textit{\% Env.} column denotes the percentage of environments in which the augmentation made an improvement, and the \textit{Avg. \% Imp.} column displays the average percentage improvement over the unaugmented baseline and standard deviation across environments.}
    \begin{tabular}{llcrr}
\toprule
    &                                     &      &  \% of Envs. &        Avg. \% Imp. \\
Algorithm & Noise Type & Noise Rate ($p$) &              &                     \\
\midrule
PPO & RandomEarlyTermination (1, 100) & 0.20 &            0 &    -28.6 $\pm$ 21.1 \\
    & RandomEarlyTermination (50) & 0.05 &            0 &    -17.2 $\pm$ 11.2 \\
    &                                     & 0.20 &            0 &    -27.8 $\pm$ 25.2 \\
    & RandomMixupObservation (0.0) & 1.00 &            0 &    -76.4 $\pm$ 39.2 \\
    & RandomNormalNoisyObservation (1.0) & 0.10 &            0 &    -29.8 $\pm$ 21.7 \\
    &                                     & 0.50 &            0 &    -48.2 $\pm$ 25.8 \\
    &                                     & 1.00 &            0 &    -55.1 $\pm$ 28.3 \\
    & RandomNormalNoisyReward (1.0) & 0.50 &            0 &    -14.7 $\pm$ 16.5 \\
    & RandomUniformScaleObservation (0.5, 1.5) & 0.50 &            0 &    -24.7 $\pm$ 20.0 \\
    &                                     & 1.00 &            0 &    -15.5 $\pm$ 14.2 \\
SAC & RandomDropoutObservation (0.1) & 0.50 &            0 &    -99.2 $\pm$ 60.9 \\
    &                                     & 1.00 &            0 &     -86.3 $\pm$ 4.8 \\
    & RandomEarlyTermination (1, 100) & 0.20 &            0 &    -26.9 $\pm$ 15.9 \\
    &                                     & 0.50 &            0 &    -55.4 $\pm$ 47.1 \\
    &                                     & 1.00 &            0 &    -98.7 $\pm$ 28.1 \\
    & RandomEarlyTermination (50) & 0.50 &            0 &    -58.0 $\pm$ 50.1 \\
    &                                     & 1.00 &            0 &   -108.2 $\pm$ 42.2 \\
    & RandomMixupObservation (0.0) & 0.20 &            0 &    -22.0 $\pm$ 42.7 \\
    &                                     & 1.00 &            0 &  -323.9 $\pm$ 501.6 \\
    & RandomNormalNoisyObservation (1.0) & 0.05 &            0 &    -50.0 $\pm$ 34.6 \\
    &                                     & 0.10 &            0 &    -75.0 $\pm$ 15.4 \\
    &                                     & 0.20 &            0 &     -90.5 $\pm$ 5.9 \\
    &                                     & 0.50 &            0 &    -98.1 $\pm$ 10.7 \\
    &                                     & 1.00 &            0 &     -96.9 $\pm$ 5.4 \\
TD3 & RandomDropoutObservation (0.1) & 0.05 &            0 &    -30.0 $\pm$ 17.6 \\
    &                                     & 0.10 &            0 &    -43.8 $\pm$ 25.4 \\
    &                                     & 0.20 &            0 &    -58.0 $\pm$ 17.8 \\
    &                                     & 0.50 &            0 &    -74.7 $\pm$ 16.4 \\
    &                                     & 1.00 &            0 &    -80.1 $\pm$ 14.6 \\
    & RandomEarlyTermination (1, 100) & 0.20 &            0 &    -34.6 $\pm$ 15.0 \\
    &                                     & 0.50 &            0 &    -51.9 $\pm$ 31.8 \\
    &                                     & 1.00 &            0 &    -98.4 $\pm$ 28.7 \\
    & RandomEarlyTermination (50) & 0.10 &            0 &    -18.3 $\pm$ 21.3 \\
    &                                     & 0.20 &            0 &    -40.0 $\pm$ 23.1 \\
    &                                     & 0.50 &            0 &    -53.8 $\pm$ 43.8 \\
    &                                     & 1.00 &            0 &   -110.1 $\pm$ 45.0 \\
    & RandomMixupObservation (0.0) & 0.20 &            0 &    -19.3 $\pm$ 17.7 \\
    &                                     & 0.50 &            0 &    -61.0 $\pm$ 59.4 \\
    &                                     & 1.00 &            0 &  -269.8 $\pm$ 378.3 \\
    & RandomMixupObservation (0.5) & 1.00 &            0 &    -36.0 $\pm$ 30.3 \\
    & RandomNormalNoisyObservation (1.0) & 0.01 &            0 &    -33.3 $\pm$ 15.1 \\
    &                                     & 0.05 &            0 &    -55.6 $\pm$ 29.6 \\
    &                                     & 0.10 &            0 &    -73.0 $\pm$ 20.6 \\
    &                                     & 0.20 &            0 &    -83.4 $\pm$ 15.4 \\
    &                                     & 0.50 &            0 &     -94.4 $\pm$ 4.9 \\
    &                                     & 1.00 &            0 &     -92.3 $\pm$ 7.0 \\
    & RandomNormalNoisyReward (0.001) & 0.10 &            0 &    -34.3 $\pm$ 46.7 \\
    & RandomNormalNoisyReward (1.0) & 0.20 &            0 &      -7.6 $\pm$ 6.2 \\
    & RandomUniformScaleReward (0.5, 1.5) & 0.01 &            0 &      -9.8 $\pm$ 6.9 \\
\bottomrule
\end{tabular}

  \label{table:results-count-bad}
\end{table*}

\subsection{Hyperparameter Choice}
\label{appendix:ablations}

\begin{figure}[ht!]%
    \centering
    \subfloat[\centering \noisetype{RandomUniformScaleReward}]{{\includegraphics[width=1.0\linewidth]{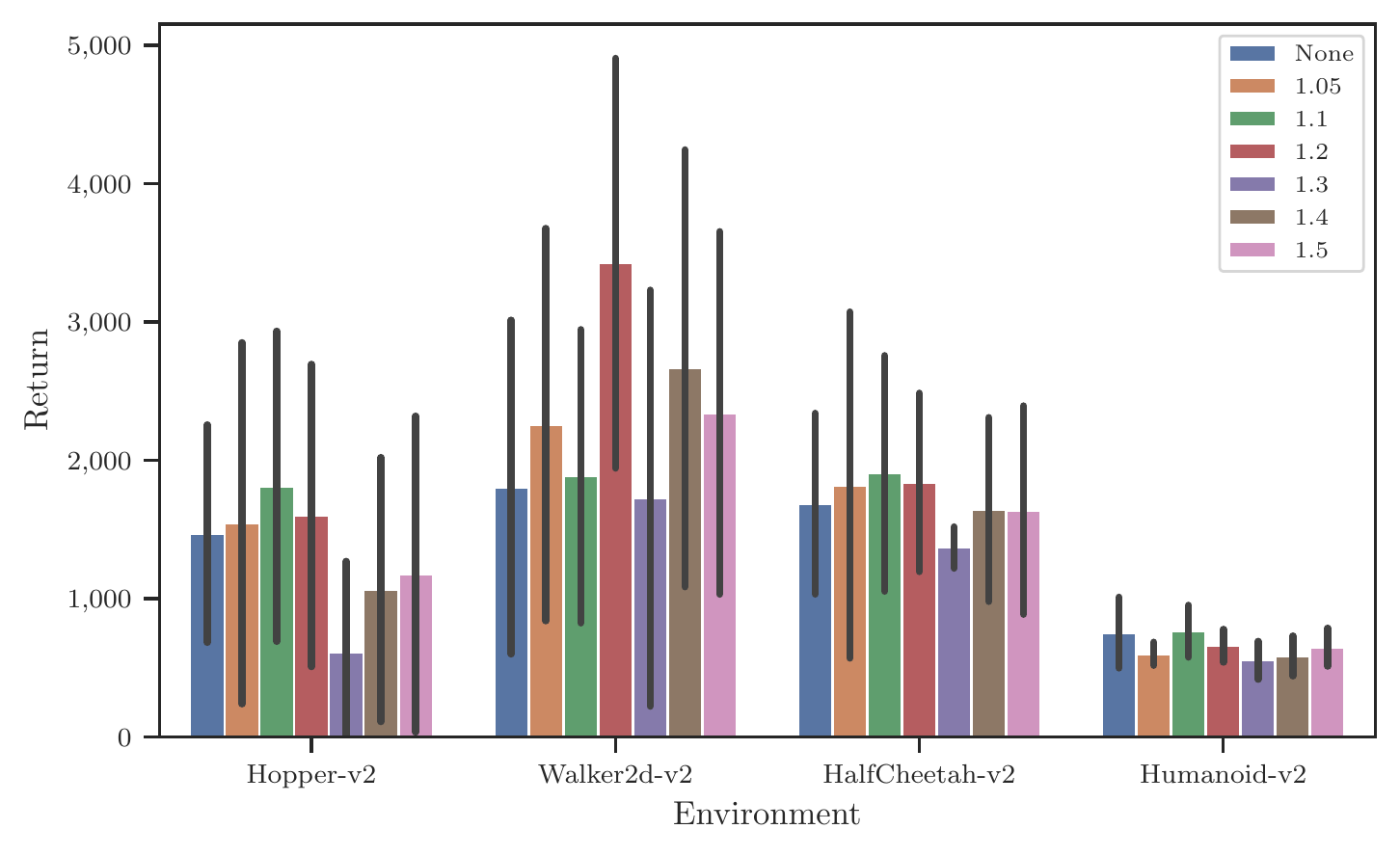} }}%
    \qquad
    \subfloat[\centering \noisetype{RandomUniformNoisyObservation}]{{\includegraphics[width=1.0\linewidth]{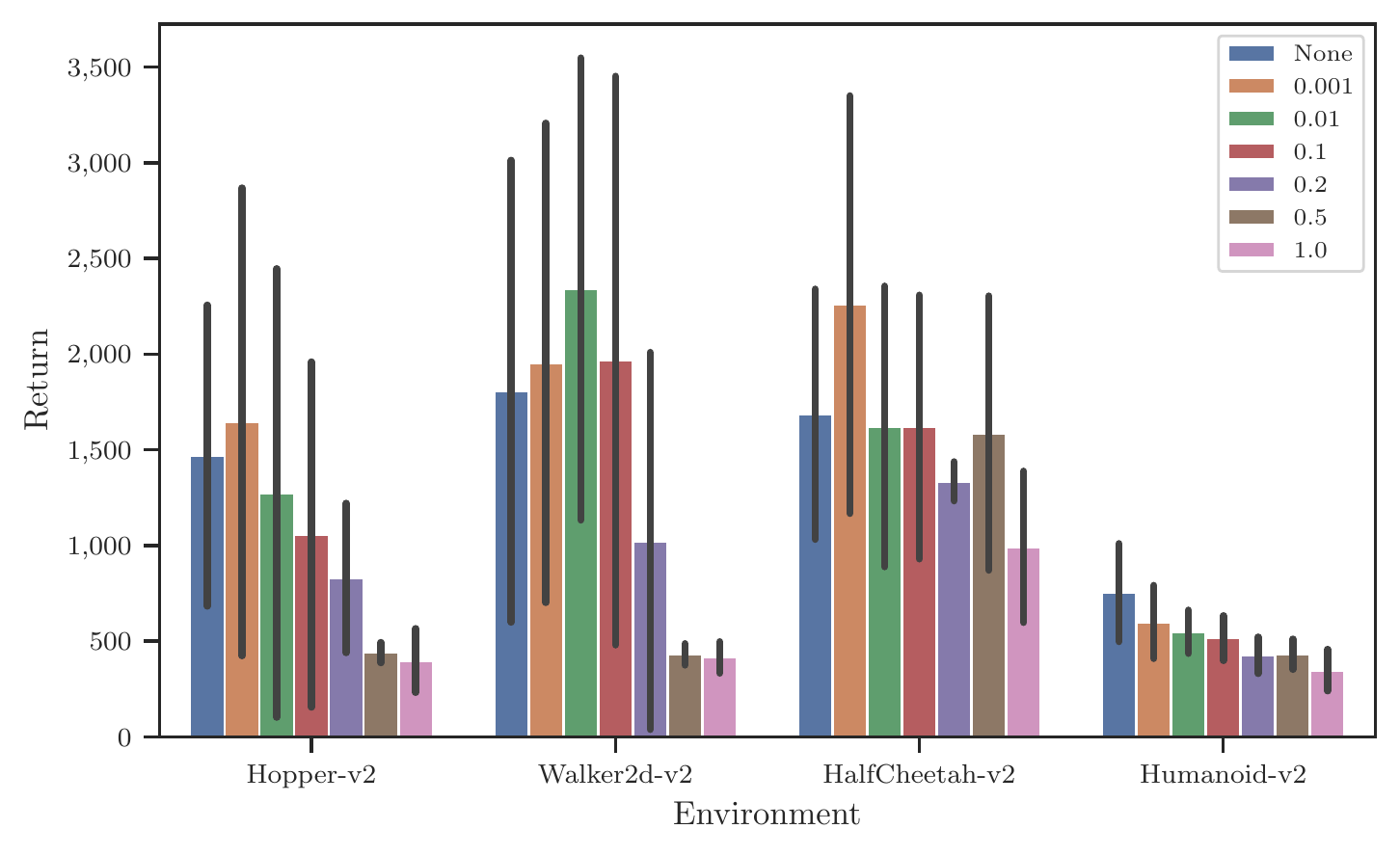} }}%
    \qquad
    \caption{Average episodic return on the unaugmented test environments across ten seeds using the PPO algorithm with different hyperparameter values for the \noisetype{RandomUniformScaleReward} and \noisetype{RandomUniformNoisyObservation} wrappers.  
For \noisetype{RandomUniformScaleReward}, the bars represent different values of $\beta$ with $\alpha = 2 - \beta$ and $p = 0.05$. 
For \noisetype{RandomUniformNoisyObservation}, the bars represent different values of $\beta$ with $\alpha = - \beta$ and $p = 1.0$. 
\noisetype{None} represents no augmentation during training.
}

    \label{fig:ppo-ablations}%
\end{figure}

Figure \ref{fig:ppo-ablations} displays the effect of hyperparameter choice for two different noise augmentation techniques on performance.

\subsection{Full Results}
\label{appendix:results}

We provide our full set of final results for the SAC, TD3, and PPO algorithms across the five MuJoCo environments in Tables \ref{table:results-summary-SAC}, \ref{table:results-summary-TD3}, and \ref{table:results-summary-PPO}.

\clearpage
\onecolumn

\small
\begin{longtable}[ht!]{rrrrrrr}
\caption{SAC average episodic returns with standard deviations after 1M steps, evaluated using 5 seeds on the test environments. Bold values indicate enhanced performance. The three best-performing augmentations per environment are highlighted in blue, while the three least effective are highlighted in red.}
\label{table:results-summary-SAC}\\
\toprule
                                    &      &                                HalfCheetah-v2 &                                    Hopper-v2 &                                  Humanoid-v2 &                                Pusher-v2 &                                  Walker2d-v2 \\
Noise Type & Noise Rate ($p$) &                                               &                                              &                                              &                                          &                                              \\
\midrule
\endfirsthead
\caption[]{SAC average episodic returns with standard deviations after 1M steps, evaluated using 5 seeds on the test environments. Bold values indicate enhanced performance. The three best-performing augmentations per environment are highlighted in blue, while the three least effective are highlighted in red.} \\
\toprule
                                    &      &                                HalfCheetah-v2 &                                    Hopper-v2 &                                  Humanoid-v2 &                                Pusher-v2 &                                  Walker2d-v2 \\
Noise Type & Noise Rate ($p$) &                                               &                                              &                                              &                                          &                                              \\
\midrule
\endhead
\midrule
\multicolumn{7}{r}{{Continued on next page}} \\
\midrule
\endfoot

\bottomrule
\endlastfoot
None & 0.00 &                             9,515 $\pm$ 1,707 &                              2,928 $\pm$ 957 &                            4,072 $\pm$ 1,913 &                              -23 $\pm$ 2 &                              4,767 $\pm$ 441 \\
RandomDropoutObservation (0.1) & 0.01 &                   \textbf{10,113 $\pm$ 1,430} &                     \textbf{3,389 $\pm$ 604} &                     \textbf{4,901 $\pm$ 923} &                     \textbf{-22 $\pm$ 2} &                            3,582 $\pm$ 2,021 \\
                                    & 0.05 &                             8,546 $\pm$ 1,579 &                            2,030 $\pm$ 1,301 &                      \textbf{5,313 $\pm$ 73} &                              -28 $\pm$ 5 &                            3,408 $\pm$ 1,294 \\
                                    & 0.10 &                             6,689 $\pm$ 1,624 &                     \textbf{3,257 $\pm$ 169} &                     \textbf{5,416 $\pm$ 178} &                              -30 $\pm$ 8 &                            2,906 $\pm$ 1,111 \\
                                    & 0.20 &                             4,506 $\pm$ 1,602 &                            2,387 $\pm$ 1,127 &                   \textbf{4,687 $\pm$ 1,392} &                             -66 $\pm$ 56 &                              1,613 $\pm$ 850 \\
                                    & 0.50 &                               1,736 $\pm$ 418 &                            1,056 $\pm$ 1,132 &                              1,357 $\pm$ 421 &          \cellcolor{pink!30}-70 $\pm$ 47 &                              1,133 $\pm$ 585 \\
                                    & 1.00 &                                 742 $\pm$ 678 &                                497 $\pm$ 214 &                                625 $\pm$ 103 &                              -41 $\pm$ 4 &                                450 $\pm$ 132 \\
\end{longtable}

\clearpage
\begin{longtable}[ht!]{rrrrrrr}
\caption{TD3 average episodic returns with standard deviations after 1M steps, evaluated using 5 seeds on the test environments. Bold values indicate enhanced performance. The three best-performing augmentations per environment are highlighted in blue, while the three least effective are highlighted in red.}
\label{table:results-summary-TD3}\\
\toprule
                                    &      &                                HalfCheetah-v2 &                                    Hopper-v2 &                                  Humanoid-v2 &                                Pusher-v2 &                                  Walker2d-v2 \\
Noise Type & Noise Rate ($p$) &                                               &                                              &                                              &                                          &                                              \\
\midrule
\endfirsthead
\caption[]{TD3 average episodic returns with standard deviations after 1M steps, evaluated using 5 seeds on the test environments. Bold values indicate enhanced performance. The three best-performing augmentations per environment are highlighted in blue, while the three least effective are highlighted in red.} \\
\toprule
                                    &      &                                HalfCheetah-v2 &                                    Hopper-v2 &                                  Humanoid-v2 &                                Pusher-v2 &                                  Walker2d-v2 \\
Noise Type & Noise Rate ($p$) &                                               &                                              &                                              &                                          &                                              \\
\midrule
\endhead
\midrule
\multicolumn{7}{r}{{Continued on next page}} \\
\midrule
\endfoot

\bottomrule
\endlastfoot
None & 0.00 &                              10,124 $\pm$ 396 &                              2,935 $\pm$ 906 &                            4,801 $\pm$ 1,258 &                              -30 $\pm$ 5 &                              4,016 $\pm$ 415 \\
RandomDropoutObservation (0.1) & 0.01 &                               8,322 $\pm$ 652 &                      \textbf{3,350 $\pm$ 91} &                            4,235 $\pm$ 2,256 &                              -38 $\pm$ 7 &                            2,960 $\pm$ 1,802 \\
                                    & 0.05 &                             7,337 $\pm$ 1,311 &                            1,319 $\pm$ 1,063 &                              4,524 $\pm$ 826 &                              -38 $\pm$ 8 &                            2,665 $\pm$ 1,158 \\
                                    & 0.10 &                             3,955 $\pm$ 2,626 &                            1,177 $\pm$ 1,062 &                              4,707 $\pm$ 725 &                              -40 $\pm$ 6 &                            1,644 $\pm$ 1,307 \\
                                    & 0.20 &                               4,638 $\pm$ 832 &                                537 $\pm$ 395 &                            2,561 $\pm$ 1,670 &                              -41 $\pm$ 5 &                            1,203 $\pm$ 1,052 \\
                                    & 0.50 &                               2,080 $\pm$ 892 &                                 330 $\pm$ 63 &                              1,186 $\pm$ 870 &                              -43 $\pm$ 6 &                                675 $\pm$ 176 \\
                                    & 1.00 &                               1,200 $\pm$ 462 &                                 295 $\pm$ 19 &                                897 $\pm$ 202 &                              -46 $\pm$ 5 &                                536 $\pm$ 139 \\
\end{longtable}

\clearpage
\begin{longtable}[ht!]{rrrrrrr}
\caption{PPO average episodic returns with standard deviations after 1M steps, evaluated using 10 seeds on the test environments. Bold values indicate enhanced performance. The three best-performing augmentations per environment are highlighted in blue, while the three least effective are highlighted in red.}
\label{table:results-summary-PPO}\\
\toprule
                                    &      &                                 HalfCheetah-v2 &                                      Hopper-v2 &                                Humanoid-v2 &                                 Pusher-v2 &                                    Walker2d-v2 \\
Noise Type & Noise Rate ($p$) &                                                &                                                &                                            &                                           &                                                \\
\midrule
\endfirsthead
\caption[]{PPO average episodic returns with standard deviations after 1M steps, evaluated using 10 seeds on the test environments. Bold values indicate enhanced performance. The three best-performing augmentations per environment are highlighted in blue, while the three least effective are highlighted in red.} \\
\toprule
                                    &      &                                 HalfCheetah-v2 &                                      Hopper-v2 &                                Humanoid-v2 &                                 Pusher-v2 &                                    Walker2d-v2 \\
Noise Type & Noise Rate ($p$) &                                                &                                                &                                            &                                           &                                                \\
\midrule
\endhead
\midrule
\multicolumn{7}{r}{{Continued on next page}} \\
\midrule
\endfoot

\bottomrule
\endlastfoot
None & 0.00 &                                1,686 $\pm$ 653 &                                1,470 $\pm$ 786 &                              754 $\pm$ 256 &                              -58 $\pm$ 14 &                              1,806 $\pm$ 1,207 \\
RandomDropoutObservation (0.1) & 0.01 &                                1,471 $\pm$ 506 &                                  689 $\pm$ 956 &                              707 $\pm$ 299 &                     \textbf{-54 $\pm$ 12} &                              1,174 $\pm$ 1,096 \\
                                    & 0.05 &                                1,681 $\pm$ 794 &                                  953 $\pm$ 716 &                              689 $\pm$ 107 &                              -62 $\pm$ 13 &                     \textbf{2,258 $\pm$ 1,474} \\
                                    & 0.10 &                     \textbf{1,882 $\pm$ 1,039} &                                  868 $\pm$ 863 &                              550 $\pm$ 120 &                     \textbf{-50 $\pm$ 12} &                     \textbf{1,849 $\pm$ 1,268} \\
                                    & 0.20 &                       \textbf{1,786 $\pm$ 867} &                                  382 $\pm$ 801 &                               519 $\pm$ 70 &                     \textbf{-57 $\pm$ 11} &                              1,555 $\pm$ 1,011 \\
                                    & 0.50 &                       \textbf{2,106 $\pm$ 948} &                                  283 $\pm$ 408 &                              579 $\pm$ 148 &                     \textbf{-56 $\pm$ 11} &                                  836 $\pm$ 908 \\
                                    & 1.00 &                                1,619 $\pm$ 598 &                                  705 $\pm$ 922 &                              494 $\pm$ 104 &                      \textbf{-57 $\pm$ 9} &                                  492 $\pm$ 424 \\
\end{longtable}

\normalsize
\clearpage
\twocolumn

\end{document}